\pdfoutput=1

\documentclass[11pt]{article}
\usepackage{booktabs}
\usepackage{adjustbox}
\usepackage{graphicx}
\usepackage{booktabs} 

\usepackage[final]{coling}

\usepackage{times}
\usepackage{latexsym}
\usepackage{cleveref}

\usepackage[T1]{fontenc}

\usepackage[utf8]{inputenc}

\usepackage{microtype}

\usepackage{inconsolata}

\usepackage{graphicx}

\title{Investigating the Impact of Language-Adaptive Fine-Tuning on Sentiment Analysis in Hausa Language Using AfriBERTa}

\author{Sani Abdullahi Sani\thanks{Correspondance to 2770930@students.wits.ac.za} \\ $^1$School of Computer Science and Applied Mathematics \\ University of the Witwatersrand, Johannesburg \\ 
         \AND
         \hspace{-2cm} Shamsuddeen Hassan Muhammad \\ \hspace{-2cm} Department of Computing\\ \hspace{-2cm} Imperial College London \\ 
        \And
        \hspace{2cm} Devon Jarvis \\ \hspace{2cm}School of Computer Science and Applied Mathematics \\ \hspace{2cm}University of the Witwatersrand, Johannesburg \\}

\begin{document}
\maketitle
\begin{abstract}
\looseness=-1
Sentiment analysis (SA) plays a vital role in Natural Language Processing (NLP) by ~identifying sentiments expressed in text. Although significant advances have been made in SA for widely spoken languages, low-resource languages such as Hausa face unique challenges, primarily due to a lack of digital resources. This study investigates the effectiveness of Language-Adaptive Fine-Tuning (LAFT) to improve SA performance in Hausa. We first curate a diverse, unlabeled corpus to expand the model's linguistic capabilities, followed by applying LAFT to adapt AfriBERTa specifically to the nuances of the Hausa language. The adapted model is then fine-tuned on the labeled NaijaSenti sentiment dataset to evaluate its performance.  
Our findings demonstrate that LAFT gives modest improvements, which may be attributed to the use of formal Hausa text rather than informal social media data. Nevertheless, the pre-trained AfriBERTa model significantly outperformed models not specifically trained on Hausa, highlighting the importance of using pre-trained models in low-resource contexts. This research emphasizes the necessity for diverse data sources to advance NLP applications for low-resource African languages. We published the code and the dataset to encourage further research and facilitate reproducibility in low-resource NLP here \href{https://github.com/Sani-Abdullahi-Sani/Natural-Language-Processing/blob/main/Sentiment%20Analysis%20for%20Low%20Resource%20African%20Languages/README.md}{SA for LowRes Language}

\end{abstract}

\section{Introduction}
\looseness=-1
Sentiment analysis (SA) is a vital task in natural language processing (NLP) aimed at identifying and categorizing opinions expressed in text \citep{Pang2007}. Although considerable progress has been made in this field, especially for widely spoken languages such as English \citep{Yimam2020}, the same cannot be said for many low-resource languages, such as Hausa \citep{Nasim2020}. Hausa is a Chadic language spoken primarily by Hausa people in the northern regions of Nigeria, Ghana, Cameroon, Benin and Togo, as well as the southern areas of Niger and Chad, with notable minority communities in Ivory Coast \citep{Wolff2024, Worldmapper, Eberhard2024}. Approximately 54 million people are estimated to speak it as their first language, while around 34 million use it as a second language, resulting in a total of about 88 million Hausa speakers \citep{Eberhard2024}. It has limited digital resources, which present challenges for NLP research, including SA \citep{Joshi2020}.

\looseness=-1
Recent advancements in pre-trained large language models (LLMs) have enabled the use of transfer learning to address challenges in NLP for low-resource languages. For example, multilingual models like BERT (Bidirectional Encoder Representations from Transformers) have shown strong performance in various NLP tasks \citep{Devlin20194171}, but often struggle with low-resource languages due to limited data and linguistic diversity \citep{Alabi20224336}. Language-adaptive fine-tuning (LAFT) has emerged as a promising approach to improve the handling of language-specific nuances in these models and improve performance in tasks such as SA, especially for underrepresented languages \citep{Pfeiffer2020}. In this study, we investigate the impact of LAFT on SA in Hausa using pre-trained LLM. We can summarize our main contributions as follows. 

\begin{enumerate}
\itemsep0em 
\item We curate a large, diverse unlabelled Hausa corpus to enrich the language's contextual and linguistic representation.
\item We show that while modest, LAFT results in a slight improvement in performance, with our model outperforming other models evaluated using the NaijaSenti dataset\footnote{The dataset and code is available at \url{https://github.com/Sani-Abdullahi-Sani/Natural-Language-Processing/tree/main}.}.

\end{enumerate}

\section{Related Work}
\label{sec:related work}

Language-Adaptive Fine-Tuning (LAFT) has demonstrated its effectiveness in enhancing sentiment analysis (SA) performance in African languages \citep{muhammad2022naijasenti}. For example, fine-tuning multilingual pre-trained language models like AfriBERTa on monolingual texts of African languages significantly improves sentiment classification tasks \citep{Alabi20224336, Wang2023488, raychawdhary-etal-2023-seals}.

AfriBERTa, introduced by \citep{Ogueji2021116}, represents a notable advancement in multilingual language modeling for African languages. It employs the Transformer architecture, leveraging the standard masked language modeling (MLM) objective for pretraining. The model is available in two configurations: a small version with approximately 97 million parameters and a large version with around 126 million parameters. This flexibility allows it to cater to varying computational resource constraints while retaining its utility for African languages.

\looseness=-1
Pre-trained on 11 African languages, AfriBERTa’s training datasets were aggregated from BBC news websites and Common Crawl, totaling less than 1 GB of data and comprising 108.8 million tokens \citep{adebara-etal-2023-serengeti}. Although the dataset size is relatively small compared to those used for other popular language models, AfriBERTa effectively captures the nuances of African languages, which is reflected in its performance on downstream NLP tasks \citep{raychawdhary-etal-2023-seals}.

\looseness=-1
AfriBERTa has been effectively utilized for SA in African languages such as Hausa and Igbo. In a study focusing on the AfriSenti-SemEval 2023 Shared Task 12, AfriBERTa was trained on annotated Twitter datasets for these languages. The model achieved impressive F1 scores of $80.85\%$ for Hausa and $80.82\%$ for Igbo, demonstrating its capability in handling sentiment classification tasks in low-resource languages \citep{raychawdhary-etal-2023-seals}.

AfriBERTa, when compared to other models like XLM-R \citep{Conneau20208440} and mBERT \citep{Devlin20194171}, has shown competitive performance. For instance, in a multilingual adaptive fine-tuning approach, AfriBERTa and XLM-R were evaluated on tasks including sentiment classification, and the results were comparable to individual language adaptations while requiring less disk space \citep{Alabi20224336}.

Another study highlighted that mBERT outperformed other models like Roberta and XLM-R in Hausa sentiment analysis, achieving the highest accuracy and F1-score of 0.73\% \citep{Yusuf202313}. However, AfriBERTa's specialization for African languages provides a significant advantage in cross-lingual transfer learning \citep{Alabi20224336}

\looseness=-1
Although multilingual fine-tuning can facilitate cross-lingual transfer learning, monolingual fine-tuning often gives superior results for specific languages. For instance, \citep{Rønningstad20231054} demonstrates that monolingual fine-tuning on datasets with thousands of samples produces optimal results. Moreover, combining language-adaptive and task-adaptive pretraining on African texts, along with careful source language selection, can lead to remarkable performance improvements. This approach minimizes harmful interference from dissimilar languages and enhances outcomes in multilingual and cross-lingual contexts \citep{Wang2023488}. Systems utilizing LAFT have achieved high rankings in shared tasks, demonstrating substantial improvements in weighted F1 scores and other performance metrics \citep{Wang2023488, Nzeyimana2023718}.

However, building reliable SA systems for low-resource African languages remains challenging due to the limited availability of training data \citep{Alabi20224336, Wang2023488}. Despite the promising results of LAFT and the benefits of monolingual fine-tuning, the scarcity of large high-quality datasets for low-resource African languages, such as Hausa, poses a significant challenge. Therefore, this study aims to contribute to the growing body of knowledge on SA for African languages by providing insights into the advantages of LAFT strategies in relation to Hausa's linguistic characteristics and availability of data.

\section{Methodology}
\label{sec:methodology}

\subsection{Conceptual Framework}
\begin{figure}[t]
  \includegraphics[width=0.9\linewidth]{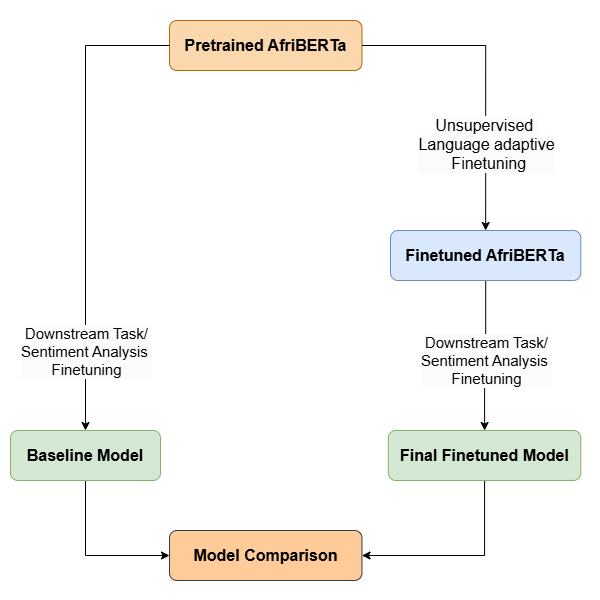}
  \caption{Experimental Overview: Assessing the Impact of the Intermediate LAFT in a Two-Phase Method for Hausa Sentiment Analysis}
\label{fig: overview}
\end{figure}

\looseness=-1
This study employs a two-phase approach to investigate the impact of LAFT on SA performance for Hausa language using the AfriBERTa model. Initially, a baseline model was established by fine-tuning AfriBERTa directly on Hausa sentiment analysis dataset (NaijaSenti), allowing us to assess its performance. Concurrently, LAFT was conducted on unlabelled data, enabling it to further adapt to the linguistic characteristics and nuances of Hausa, resulting in a refined model. The refined model is then saved and reloaded into the same pipeline, where it undergoes a second fine-tuning process on NaijaSenti with the same set of parameters as the baseline model. It is hypothesized that this two-stage fine-tuning method, which is depicted in Figure ~\ref{fig: overview}, would improve the model's sentiment classification performance and produce a final model that is optimal for Hausa SA.


\begin{table*}[htbp]
  \caption{Distribution of LAFT Data Sources, Including the Approximate Number of Sentences Collected and Their Respective Domains Covered}
  \centering
  \begin{adjustbox}{width=\textwidth}
  \begin{tabular}{p{4cm}p{4cm}p{5cm}}
    \toprule[1.5pt]
    \textbf{Data Source} & \textbf{No. of Data Examples} & \textbf{Domain Covered} \\
    \midrule
    Hausa Global Media & 15,000 & Business, Psychology, Healthcare, Education, Religion, Self-Awareness, Technology, Politics \\
    Hausa Novel Store  & 20,000 & Romance, Entertainment, Healthcare \\
    Scanned Literature & 5,000  & Classic Literature \\
    \bottomrule[1.5pt] 
  \end{tabular}
  \end{adjustbox}

  \label{tab: data sources}
\end{table*}

\subsection{Dataset Collection}
    
\textbf{General Fine-Tuning Dataset:}
\looseness=-1
\Cref{tab: data sources} presents the distribution of the LAFT corpus we collected for this study with their respective domain. Table ~\ref{tab:gen_finetuning_data} displays examples of this data in Hausa, the corresponding English translations, and the respective domains they originate from. We employed three distinct data collection approaches as described below:

\begin{itemize}
    \item \textbf {Hausa Global Media:}
     In collaboration with the blogging platform, we obtained a dataset of approximately 15,000 sentences, including short and long blogs, as well as books covering diverse topics such as Business, Psychology, Healthcare, Education, Religion, Self-Awareness, Technology, and Politics. We provided an incentive to the company as a token of appreciation for their contribution. 
     
    \item \textbf {Hausa Novel Store: }
    \looseness=-1
    We scraped content from Hausa novel store website \footnote{https://hausanovel.ng/} , an online store for Hausa novels, resulting in around 20,000 sentences focusing on Romance, Entertainment, and Healthcare. The content of the website is freely available on public domain.
    \item \textbf {Scanned Literature:} 
    We accessed scanned copies of classic Hausa literature, including notable titles like "Magana Jari Ce" and "Ruwan Bagaja." from archive.org website\footnote{https://archive.org/}. Using Tesseract OCR with Python, we extracted text from these scanned books, yielding approximately 5,000 sentences. The collected data was then preprocessed for further analysis.
\end{itemize}

For further details regarding the data curation ethics see Section \ref{sec:data curation ethics}.

\begin{table*}[htbp]
  \caption{Examples of LAFT Data, Their English Translations, and Respective Domains}
  \centering
  \begin{tabular}{p{5cm}p{5cm}p{2cm}}
    \hline
    \textbf{Example in Hausa} & \textbf{Translation (English)} & \textbf{Domain} \\
    \hline
    A dabi'ar dan adam ba kasafai ya fiya son canji ba & Human nature rarely likes change & Psychology \\
    \hline
    Ya sayi haja ta kasuwanci, ya sayar da rabi a hanya. & He bought stock for business, sold half on the way. & Business \\
    \hline
    Menene manufar zuwan Annabi? & What is the purpose of the Prophet's coming? & Religion \\
    \hline
  \end{tabular}
  \label{tab:gen_finetuning_data}
\end{table*}

\textbf{Downstream Task Dataset:}
For the downstream task, we used the NaijaSenti dataset by \citep{Muhammad20232319}, which is publicly available on Hugging Face. This dataset, designed for SA on individual tweets from Twitter, has been pre-processed and annotated with sentiment labels: Neutral, Positive, and Negative. The NaijaSeni dataset serves as a benchmark for evaluating the sentiment classification performance of our model.

\subsection{Dataset Cleaning and Preprocessing}

For the LAFT Corpus preprocessing, we removed extra whitespaces, trimmed leading and trailing spaces, and split the text into sentences using sentence-ending punctuation (e.g., periods, exclamation marks, question marks). The NaijaSenti dataset is already cleaned, requiring no additional preprocessing.

\subsection{Tokenization}
We employed the AutoTokenizer from the Hugging Face library \citep{ThomasWolfetal2019} for the AfriBERTa model \citep{Ogueji2021116}, utilizing the SentencePiece algorithm \citep{DTakuKudo2018} for subword tokenization. This method effectively handles rare words and morphologically rich languages by breaking down text into smaller subword units, ensuring meaningful representation of out-of-vocabulary words. We maintained the maximum sequence length of 512 tokens, standardizing input data by truncating longer sequences and padding shorter ones by a special padding token '0'. This preprocessing step is crucial for converting raw text into numerical tokens that the model can process efficiently, maintaining a consistent input format for the SA tasks.

\subsection{Dataset Split}
The LAFT and downstream task datasets were divided into training, validation, and testing sets using a 70:10:20 ratio. This resulted in 30,866 training, 4,412 validation, and 8,826 test examples for the LAFT dataset, and 18,989 training, 2,714 validation, and 5,427 test examples for the downstream SA task, as shown in Table  ~\ref{tab:dataset_splits}.

\begin{table}[htbp]  
  \caption{Dataset splits for LAFT and sentiment analysis}
  \centering
  \begin{tabular}{cccc}
    \hline
    \textbf{Dataset} & \textbf{Train} & \textbf{Val} & \textbf{Test} \\
    \hline
    LAFT Corpus & 30,866 & 4,412 & 8,826 \\
    NaijaSenti (Hausa) & 18,989 & 2,714 & 5,427 \\
    \hline
  \end{tabular}
  \label{tab:dataset_splits}
\end{table}


\subsection{Model Selection}
\looseness=-1
We selected the AfriBERTa small model \citep{Ogueji2021116} for our experiments due to its pre-training on African languages, which aligns with the objectives of our study. AfriBERTa is a multilingual language model with approximately 97 million parameters, 4 layers, 6 attention heads, 768 hidden units, and a feed-forward size of 3072. It was pre-trained on 11 African languages—including Afaan Oromoo, Amharic, Gahuza, Hausa, Igbo, Nigerian Pidgin, Somali, Swahili, Tigrinya, and Yorùbá. AfriBERTa’s multilingual capabilities enable it to capture complex linguistic patterns and perform well on tasks such as text classification and Named Entity Recognition across diverse African languages.

Our motivation is largely driven by our computational constraints. This smaller version provides an efficient balance between performance and resource requirements while retaining the linguistic advantages of its larger counterpart, making it suitable for our task.

\subsection{Model Evaluation}
\looseness=-1
We evaluate model performance using accuracy, precision, recall, and F1-score. We also used the training and validation loss to monitor the model's learning process, particularly during training, to have an idea about model complexity.\footnote{We conducted experiments using Google Colab Pro environment with a T4 GPU.}

\subsection{Model Training and Optimization}

We employed the Hugging Face Transformers \texttt{Trainer} API, utilizing the AdamW optimizer with weight decay set to 0.01 to control overfitting. A batch size of 8 was used consistently across training and evaluation phases. For both the LAFT phase and the downstream SA task, we initially set the learning rate at \(2 \times 10^{-5}\). Observations of early overfitting, as indicated by a rise in validation loss after the first epoch, prompted a reduction to \(1 \times 10^{-5}\), resulting in stable convergence and improved performance.

In terms of epochs, we determined through experimentation that 5 epochs were optimal for the LAFT phase, while 3 epochs provided a balance of generalization and efficiency in the SA task. Evaluation was conducted at the end of each epoch, with the best-performing model retained based on validation metrics.

In comparison, AfriBERTa Large is known in the literature for achieving higher performance; our baseline experiment confirmed this with an F1 score of 0.79 and an evaluation loss of 0.95. However, it required significantly more computational resources (874.6 seconds of train runtime) compared to AfriBERTa Small, which achieved an F1 score of 0.77 with lower evaluation loss (0.582) and faster train runtime (397.9 seconds). Given these findings, we selected AfriBERTa Small for its efficiency and near-parity in performance within our resource constraints.






\begin{table*}[ht]
  \caption{Performance metrics for downstream SA task before and after LAFT, averaged over three runs. Standard deviation is ±0.01 for all performance metrics.}
  \centering
  \begin{tabular}{lcccc}
    \toprule
    & \multicolumn{4}{c}{\textbf{Performance Metrics}} \\
    \cmidrule(lr){2-5}
    & \textbf{Accuracy (\%)} & \textbf{F1 (\%)} & \textbf{Precision (\%)} & \textbf{Recall (\%)} \\ 
    \midrule
    \textbf{Before LAFT} & & & & \\
    Training   & 77.00\scriptsize{$\pm$0.01} & 77.00\scriptsize{$\pm$0.01} & 78.00\scriptsize{$\pm$0.01} & 77.00\scriptsize{$\pm$0.01} \\ 
    Validation & 77.00\scriptsize{$\pm$0.01} & 77.00\scriptsize{$\pm$0.01} & 77.00\scriptsize{$\pm$0.01} & 77.00\scriptsize{$\pm$0.01} \\ 
    Testing    & 75.00\scriptsize{$\pm$0.01} & 75.00\scriptsize{$\pm$0.01} & 76.00\scriptsize{$\pm$0.01} & 75.00\scriptsize{$\pm$0.01} \\ 
    \cmidrule(lr){2-5}
    \textbf{After LAFT} & & & & \\
    Training   & 78.00\scriptsize{$\pm$0.01} & 78.00\scriptsize{$\pm$0.01} & 77.00\scriptsize{$\pm$0.01} & 78.00\scriptsize{$\pm$0.01} \\ 
    Validation & 78.00\scriptsize{$\pm$0.01} & 78.00\scriptsize{$\pm$0.01} & 78.00\scriptsize{$\pm$0.01} & 78.00\scriptsize{$\pm$0.01} \\ 
    Testing    & 75.00\scriptsize{$\pm$0.01} & 75.00\scriptsize{$\pm$0.01} & 76.00\scriptsize{$\pm$0.01} & 75.00\scriptsize{$\pm$0.01} \\ 
    \bottomrule
  \end{tabular}
  \label{tab:performance_metrics_combined}
\end{table*}

\section{Results}
\label{sec:results}
The results, averaged over three runs with a variation of \(\pm 0.01\), are presented across several metrics, comparing the model's performance before and after LAFT. A detailed analysis of both the baseline and LAFT models is provided below.

\subsection{Performance Metrics Before LAFT (Baseline Model)}

\begin{figure*}[htbp]
  \includegraphics[width=0.48\linewidth]{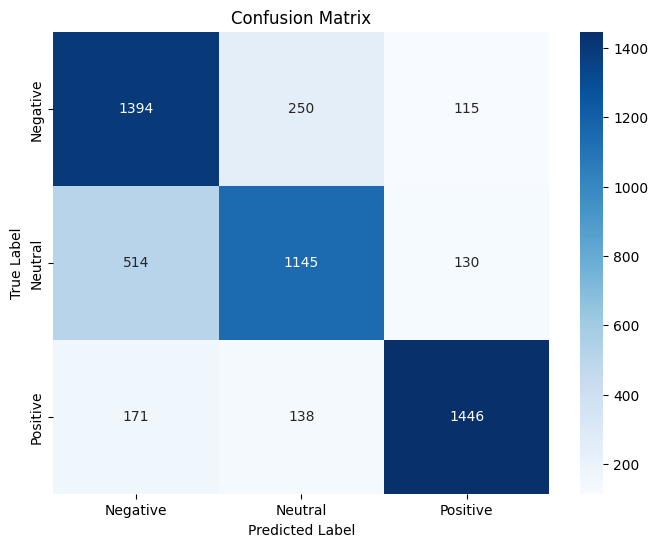} \hfill
  \includegraphics[width=0.48\linewidth]{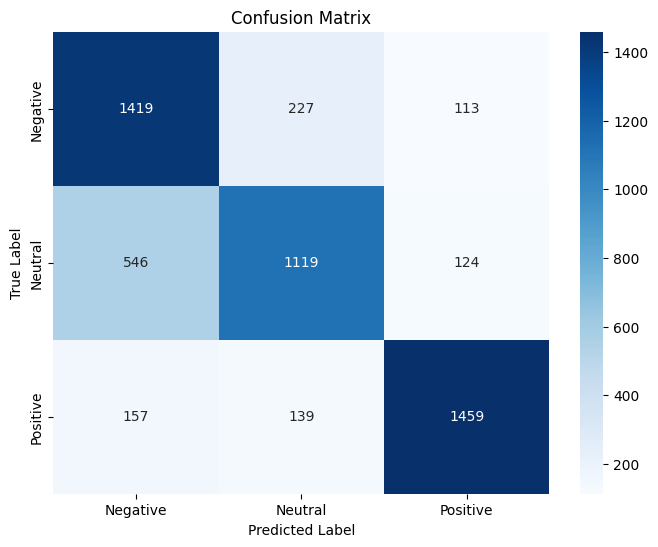}
  \caption {Confusion Matrix for Downstream Task before LAFT (Baseline Model on the left), and after LAFT (on the right)}
  \label{fig: conf_matrix for baseline and after LAFT}
\end{figure*}

 \Cref{tab:performance_metrics_combined} summarized the baseline model's performance. The model achieved a training accuracy of 77\%, consistent across training and validation, with both reaching approximately 77-78\%. Precision, Recall, and F1-Score are closely aligned, indicating balanced performance and minimal bias against specific classes. The confusion matrices in ~\Cref{fig: conf_matrix for baseline and after LAFT} confirm this, showing no significant errors in classifying Positive and Negative sentiments. However, the model tends to misclassify neutral sentiments as negative, likely due to an overlap between neutral and negative expressions in the dataset, making it challenging for the model to distinguish subtle differences.

\subsection{Performance Metrics After LAFT}

After LAFT as seen in Table \ref{tab:performance_metrics_combined}, training accuracy, F1, and Recall showed a slight improvement from 77\% to 78\%. Validation performance also increased from 77\% to 78\%, while testing accuracy remained nearly identical, with metrics ranging from 75\% to 76\%. 


\begin{table}[ht]
  \caption{Training and Validation Loss for LAFT}
  \centering
  \begin{tabular}{ccc}
    \hline
    \textbf{Epoch} & \textbf{Training Loss} & \textbf{Validation Loss} \\ 
    \hline
    1 & 3.229 & 3.035 \\ 
    2 & 3.092 & 2.957 \\ 
    3 & 3.033 & 2.907 \\ 
    4 & 2.954 & 2.887 \\ 
    5 & 2.923 & 2.890 \\ 
    \hline
  \end{tabular}
  \label{tab:laft_training_and_validation_loss}
\end{table}

\begin{figure*}[htbp]
  \includegraphics[width=0.48\linewidth]{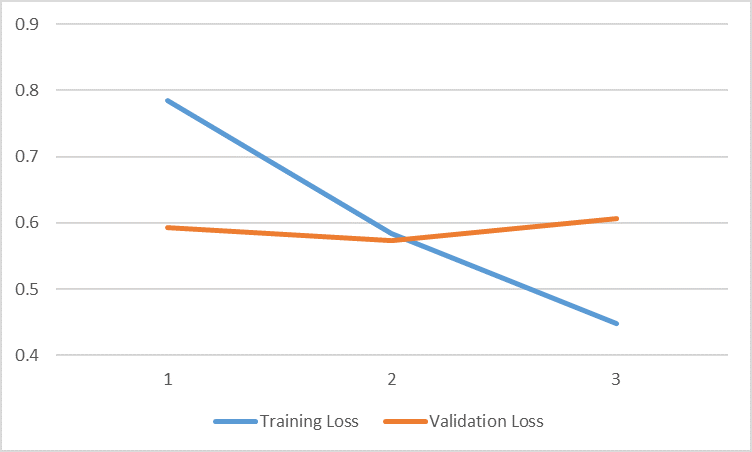} \hfill
  \includegraphics[width=0.48\linewidth]{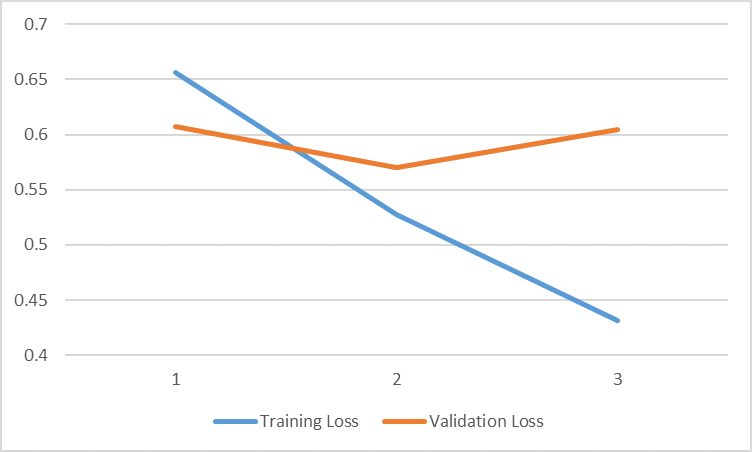}
  \caption {Training and Validation Loss for the Downstream Task before and after LAFT. The graph indicate that the model after LAFT (to the right) demonstrates effective learning, beginning with lower training loss compared to the baseline model before LAFT (to the left), highlighting the benefits of the fine-tuning process}
  \label{fig: Downstream train and val loss before and after LAFT}
\end{figure*}

Figure ~\ref{fig: Downstream train and val loss before and after LAFT} present the training and validation losses before and after LAFT, respectively. The plots indicate that the model after LAFT (to the right) consistently starts with lower training losses (approximately 0.66 compared to around 0.79 before the LAFT), suggesting better initial learning which shows that LAFT is effectively enhancing the learning of our model. In both models though, training loss steadily decreases over the epochs, demonstrating improved performance as training progresses. However, while validation loss decreases initially, it begins to rise slightly by the third epoch, suggesting potential overfitting in both models. This overfitting may be attributed to limited data availability and the lack of standardized orthographic forms in many African languages \citep{Mohamed2024207, Baguma20243}, leading to inconsistencies that hinder the model's ability to generalize effectively.

\begin{figure}[htbp]
\centerline{\includegraphics[width=\linewidth]{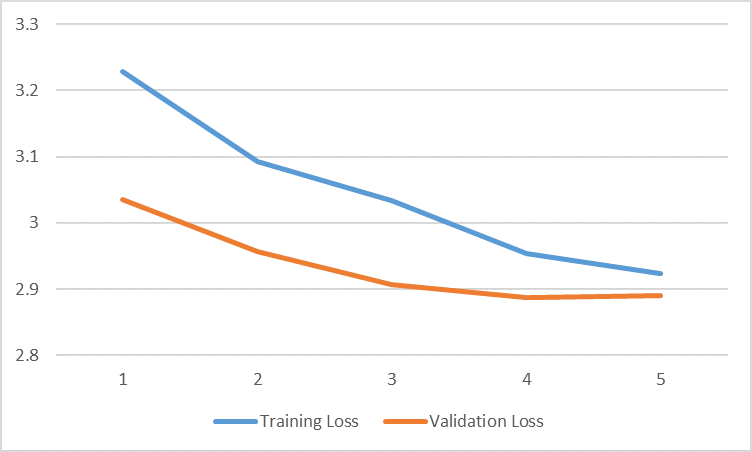}}
\caption{LAFT Training and Validation Loss curve across five epochs showing a consistent reduction, indicating effective learning. However, by the fifth epoch, the validation loss begins to rise slightly, suggesting a potential sign of overfitting}
\label{fig: LAFT train and val loss}
\end{figure}

\begin{figure*}[htbp]
  \includegraphics[width=0.48\linewidth]{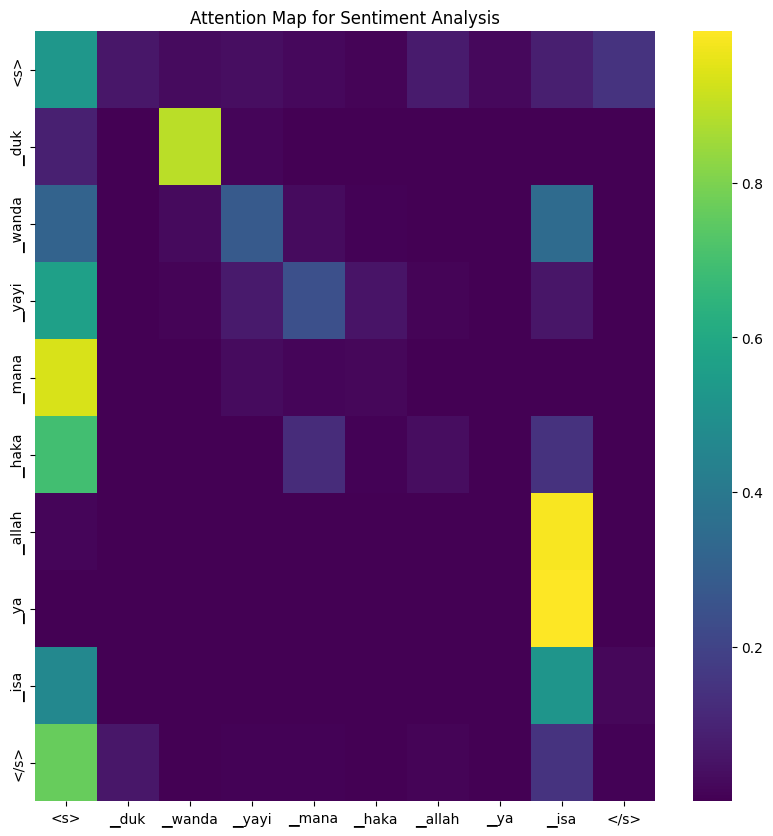} \hfill
  \includegraphics[width=0.48\linewidth]{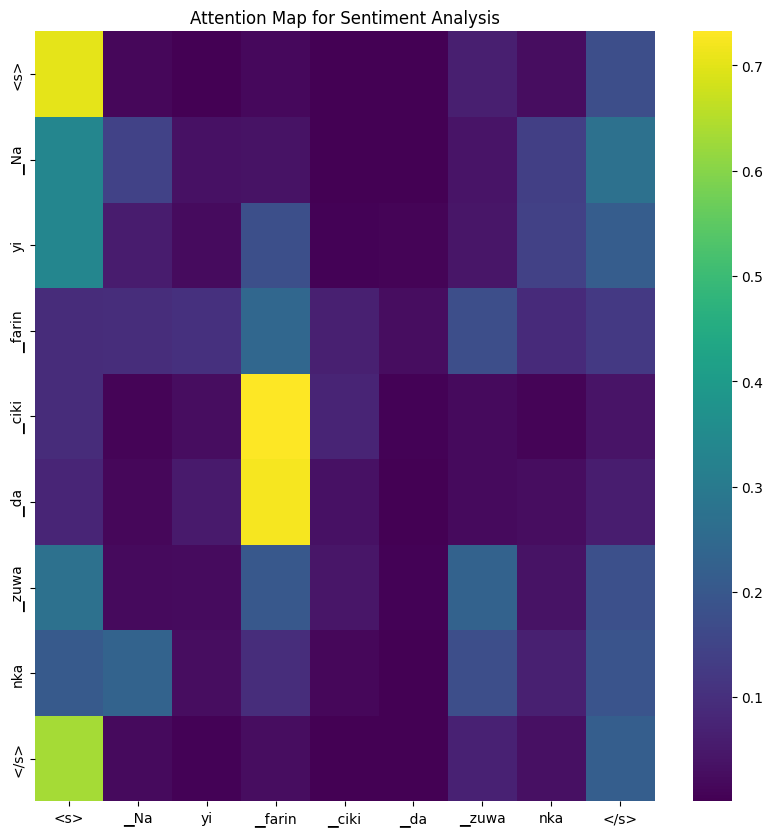}
  \caption {Attention Map Highlighting Key Phrases in Sentiment Analysis with strong focus on 'Allah ya isa' Indicating Negative Sentiment. On the right side, showing the model attending to the phrase "farin ciki" (glad) and "da" (that), demonstrating its capability to effectively capture positive sentiment in the text}
  \label{fig:attention map for positive and negative sentiment}
\end{figure*}

\looseness=-1
The attention map in Figure ~\ref{fig:attention map for positive and negative sentiment} for the sentence "duk wanda yayi mana haka allah ya isa" by the Baseline Model reveals that the model strongly focuses on the tokens  "Allah" and "ya" and "isa". This is notable because the phrase "Allah ya isa" roughly translates to "I won't forgive you" or "Allah will be the judge," which conveys a clear negative sentiment. The model's attention on this part of the sentence suggests that it is effectively identifying the most important section contributing to the overall sentiment. Since "Allah ya isa" carries the emotional weight of unforgiveness, the model’s focus here supports its prediction of negative sentiment. This alignment between attention and meaning demonstrates that the model not only makes accurate predictions but also does so in an interpretable way by zeroing in on the part of the text that holds the strongest emotional significance. Additionally, the other attention map from the model after LAFT is provided on the right in Figure ~\ref{fig:attention map for positive and negative sentiment}, which explains how the sentence "Nayi farin ciki da zuwanka," meaning "I'm glad you're here" is processed. In this case, the model attends strongly to the words "farin ciki" (Glad) and "da" (that), highlighting its ability to capture positive sentiment as well.

\section{Discussion}
\label{sec:discussion}

Despite the subtle improvements in validation metrics, our findings align with previous studies by \citep{Alabi20224336} and \citep{Wang2023488}, which demonstrate that fine-tuning a multilingual pre-trained language model (PLM) on monolingual texts enhances sentiment classification performance for African languages. 

Compared to previous SA works in Hausa, our results shows notable improvements. For instance, \citep{mubarak2024fine} achieved an accuracy of 66.0\% and an F1 score of 66.0\% with the Gemma 7B model on the NaijaSenti Hausa dataset. Our model outperforms it, highlighting the efficacy of LAFT and AfriBERTa in understanding Hausa nuances. Similarly,\citep{Kumshe2024HausaSentimentAnalysis} fine-tuned a BERT-based model on the same dataset, achieving an accuracy of 73.47\%. Our model surpasses this performance, further demonstrating that AfriBERTa's design for African languages provides a significant advantage in capturing linguistic nuances within Hausa text. However, \citep{muhammad2022naijasenti} utilized the AfriBERTa large model and achieved an accuracy of 81.2\%. Our findings with the smaller model (AfriBERTa small) still show notable competitive performance, especially considering the model size. Thus, our model performance not only validate the efficacy of the approach but also highlight the importance of using pre-trained models like AfriBERTa that already incorporate African languages, leading to improved performance on sentiment classification tasks.

\section{Conclusion}
In this study, we explored the use of LAFT for SA in Hausa, a low-resource language, leveraging AfriBERTa, which is pre-trained on African languages including Hausa. AfriBERTa’s pre-training offered a notable advantage, outperforming models not trained on Hausa by effectively capturing its linguistic nuances. Although LAFT resulted in slight performance improvements, it did not significantly exceed the baseline set by AfriBERTa’s pre-training. This limited improvement is likely due to the fine-tuning corpus, which consisted mostly of formal text, contrasting with the conversational language commonly used in sentiment tasks. Our results highlight the need for more diverse datasets that include informal and dialectal variations to boost generalization and performance. Future efforts should prioritize expanding both data sources and fine-tuning techniques to enhance NLP tasks in low-resource languages like Hausa.

\section{Limitations}
\label{sec:limitation}
\looseness=-1
While our dataset covers a broad range of topics, domains like Business, Healthcare, and Romance are overrepresented compared to others like Technology and Politics, This imbalance could affect the model’s ability to generalize effectively, potentially limiting its performance in the downstream SA tasks.

\looseness=-1
A potential reason why our LAFT approach may not have significantly improved performance could be the nature of the training corpus, which primarily consists of formal Hausa text, such as literature, rather than the informal, conversational language common on social media. Privacy policies restricted our ability to collect enough social media data, which likely impacted the model’s effectiveness in SA tasks.

Additionally, our LAFT dataset mainly represents the Kano Hausa dialect, which may cause the model to underperform with other dialects. Due to limited available data for these dialects, we could not include them in the training process, limiting the model's generalizability to other dialects.

\section{Future Work}
\label{sec:future_work}
An important direction for future research is to investigate the performance of other multilingual models such as XLM-R, AfroXLMR, and mBERT. Comparing these models' capabilities in capturing Hausa linguistic nuances could provide deeper insights into SA for low-resource languages.

\looseness=-1
Our current dataset primarily consists of formal, structured text. Future work should focus on collecting and incorporating more diverse datasets, particularly those containing less structured language from social media platforms. By introducing more conversational and informal text, we can improve the model's ability to generalize and capture the subtle sentiment variations present in everyday language.

Combining AfriBERTa with other state-of-the-art models like mBART and XLM-R could potentially enhance its performance in multilingual and cross-lingual tasks, addressing the limitations of individual models \citep{Mathur2024}

During our tokenization process, we observed that some words were broken down into subwords that might not preserve their original semantic meaning. A promising future research direction is to develop a custom tokenizer specifically trained on Hausa lexicons for SA. This approach could potentially preserve whole words and prevent unnecessary fragmentation; researchers might improve the model's sensitivity to semantic nuances, particularly in distinguishing subtle positive and neutral sentiment expressions.

\section{Ethical Considerations}
\label{sec:ethical_considerations}

\begin{enumerate}
\item {\bf Explainability and Safety}
    \par In our research, we prioritize the explainability of our model to ensure safety and trustworthiness. We visualize attention maps for specific data subsets, which illustrate how our model focuses on critical tokens during prediction.
    
\item {\bf Broader Impacts}
    \par We address both potential positive and negative societal impacts of our work.
    \begin{itemize}
            \item {\bf Positive Impacts:}
            \par Our project aims to improve sentiment         analysis for low-resource languages and       promote inclusivity in NLP.
            
            \item {\bf Negative Impacts:}
            \par We acknowledge the risk of perpetuating biases inherent in the training data.
    \end{itemize} 

\item {\bf Licensing of Existing Assets}
    \par We ensure that the creators or original owners of the assets used in our paper are properly credited.We explicitly mention the use of publicly available datasets and models, citing them appropriately.

\item {\bf Data Curation Ethics Statement}
    \label{sec:data curation ethics}
    \par In our collaboration with Hausa Global Media, we initiated discussions to explore our research project focusing on NLP and the critical need for a comprehensive Hausa language corpus. The platform expressed a strong commitment to supporting our efforts in advancing Hausa NLP research and agreed to share their dataset. To recognize the contributions of the staff involved in collating this dataset, we provided a modest incentive as a token of appreciation for their valuable work. Importantly, this incentive was carefully structured to ensure that it did not influence the integrity or objectivity of the data collection process, thereby preventing any potential bias.

    Additionally, we gathered data from publicly accessible platforms, including Hausa Novel and the Internet Archive. The content from Hausa Novel is openly available to anyone, and we made sure to collect this data in accordance with their privacy policies. For literature sourced from Internet Archive, we adhered to their established guidelines. The Internet Archive explicitly states on their website that it is a 501(c)(3) non-profit organization dedicated to building a digital library of Internet sites and other cultural artifacts in digital form, providing free access to researchers, historians, scholars, individuals with print disabilities, and the general public. We ensured strict compliance with their privacy policies and data agreements, acknowledging their significant contributions to making this data available.

    we are committed to ethical data curation practices, prioritizing transparency and integrity throughout our research process. All relevant materials can be found here: 
    \href{https://github.com/Sani-Abdullahi-Sani/Natural-Language-Processing/blob/main/Sentiment%20Analysis%20for%20Low%20Resource%20African%20Languages/README.md}{SA for LowRes Language}
    
\end{enumerate}

\section*{Acknowledgments}
This work is supported by a DeepMind scholarship to S.A.S. to pursue studies at the University of the Witwatersrand, Johannesburg. D.J. is a Google PhD Fellow and Commonwealth Scholar.

\bibliography{custom}

\newpage
\appendix

\section*{Appendix}
\section{Hyperparameters}
\label{sec:hyperparameter}

\begin{table}[ht]
    \centering
    \caption{Hyperparameters}
    \begin{tabular}{cc}
        \hline
        \textbf{Hyperparameter}            & \textbf{Value}                               \\ \hline
        Training Batch Size                & \quad 8                                       \\ \hline
        Evaluation Batch Size              & \quad 8                                       \\ \hline
        Epochs                             & \quad 3 (SA), 5 (LAFT)                       \\ \hline
        Learning Rate                      & \quad \(1 \times 10^{-5}\)                   \\ \hline
        Weight Decay                       & \quad 0.01                                    \\ \hline
        Eval Strategy                & \quad End of epoch                            \\ \hline
    \end{tabular}
    \label{tab:hyperparameters}
\end{table}

\end{document}